\documentclass[10pt,twocolumn,letterpaper]{article}

\usepackage{cvpr}              

\usepackage[dvipsnames]{xcolor}

\definecolor{cvprblue}{rgb}{0.21,0.49,0.74}
\usepackage[pagebackref,breaklinks,colorlinks,citecolor=cvprblue]{hyperref}
\usepackage{multirow}
\usepackage{graphicx}
\usepackage{listings}
\usepackage{xcolor}
\usepackage{algorithm}
\usepackage{algpseudocode}  

\def\paperID{*****} 
\def\confName{CVPR}
\def\confYear{2024}

\title{Symmetric Multi-Similarity Loss \\for EPIC-KITCHENS-100 Multi-Instance Retrieval Challenge 2024 }

\author{Xiaoqi Wang \quad Yi Wang \quad Lap-Pui Chau \vspace{0.3em} \\
{Department of Electrical and Electronic Engineering, The Hong Kong Polytechnic University}\\
{\tt\small xiaoqi.wang@connect.polyu.hk, \{yi-eie.wang, lap-pui.chau\}@polyu.edu.hk}
}

\begin{document}
\maketitle
\begin{abstract}
In this report, we present our champion solution for EPIC-KITCHENS-100 Multi-Instance Retrieval Challenge in CVPR 2024. Essentially, this challenge differs from traditional visual-text retrieval tasks by providing a correlation matrix that acts as 
 a set of soft labels for video-text clip combinations. However, existing loss functions have not fully exploited this information. Motivated by this, we propose a novel loss function, Symmetric Multi-Similarity Loss, which offers a more precise learning objective. Together with tricks and ensemble learning, the model achieves 63.76\% average mAP and 74.25\% average nDCG on the public leaderboard, demonstrating the effectiveness of our approach. Our code will be released at: \url{https://github.com/xqwang14/SMS-Loss}.
\end{abstract}    

\section{Introduction}

The goal of visual-text retrieval task is to accurately match visual content, such as images or videos, with corresponding natural language descriptions. This task is important for various applications, including  content-based image search and multimedia recommendation systems. Consequently, we have witnessed a rapid growth, with a numbers of innovative approaches being proposed to improve retrieval accuracy and efficiency \cite{ dgl, gat}. The EPICKITCHENS-100 (EK-100) Multi-Instance Retrieval Challenge \cite{ek100_1, ek100_2} is distinguished by the inclusion of a relevancy matrix that defines the relationships between verbs and nouns \cite{mi-mm}. This feature makes it more challenging for evaluating retrieval methods.

AVION \cite{AVION} provides us an ideal baseline model as they leverage the vanilla CLIP-based \cite{clip} model to achieve impressive performance with minimal computational cost. However, upon exploring existing loss functions, we found that the learning objective of the current state-of-the-art loss function, the Adaptive Max-Margin Multi-Instance (Adaptive MI-MM) Loss \cite{egovlp}, is not always correct due to the hard mining strategy employed in their algorithm. 

Specifically, the hard mining strategy deployed by the Adaptive MI-MM loss allows the dataloader to select hard positive samples where the correlation value in the relevancy matrix is lower than 1. In this situation, there is a possibility that negative pairs have a stronger relation to the corresponding textual descriptions, leading the model to optimize in the wrong direction. However, directly removing the hard mining strategy would result in a significant drop in the performance of models.

To address this issue, we propose a novel loss function called Symmetric Multi-Similarity (SMS) Loss, which is adapted from Multi-Similarity Loss \cite{ms_loss}, and offers a more precise learning objective for the EK-100 Multi-Instance Retrieval Challenge,  by symmetrically optimizing the positive and negative pairs. Concretely, we redefined the correlation between positive and negative pairs, and a relaxation factor is added to prevent the loss between similar pairs from becoming dominant. Meanwhile, to obtain better result, we introduce a simple but useful trick of horizontally flipping video frames during the inference phase and employ an ensemble of diverse models.

In experiments, we achieved the 1st place in the EK-100 Multi-Instance Challenge 2024 on the public leaderboard. Our method, which adopts the previous state-of-the-art solution as the baseline algorithm, significantly improves its performance, i.e.,  an average mAP increase from 58.42\% to 63.76\% (+5.34\%) and an average nDCG increase from 70.76\% to 74.25\% (+3.49\%).

\definecolor{codegreen}{rgb}{0,0.6,0}
\definecolor{codegray}{rgb}{0.5,0.5,0.5}
\definecolor{codepurple}{rgb}{0.58,0,0.82}
\definecolor{blackcolour}{rgb}{0.95,0.95,0.92}

\lstdefinestyle{python_style}{
    language=python,
    basicstyle=\small\ttfamily,
    keywordstyle=\color{codepurple},
    commentstyle=\color{codegreen},
    numberstyle = \color{codegray},
    morekeywords = {torch},
    morekeywords = {np},
    stringstyle=\color{red},
    breaklines=true,
    showstringspaces=false,
}
\lstset{style=python_style}

\section{Methodology}
\label{sec:formatting}
In this session, we begin by outlining the typical learning objective in deep metric learning. We then delve into the development of our SMS loss, explaining its derivation and why it is well-suited for the EK-100 Multi-Instance Retrieval Challenge.
\subsection{Preliminaries}

For a typical visual-text retrieval task, a\vspace{0.2mm} given triplet set $\mathcal{D}=\{\mathcal{V},\mathcal{T},\mathcal{C}\}$ is provided as input.\vspace{0.2mm} Here, $\mathcal{V}=\{\mathbf{v}_i\}^{\mathcal{N}_v}_{i=1}$ represents the video set \vspace{0.5mm} and $\mathcal{T}=\{\mathbf{t}_j\}^{\mathcal{N}_t}_{j=1}$ represents narration set, with ${\mathcal{N}_v}$ and ${\mathcal{N}_t}$ samples, respectively. The label $\mathcal{C}=\{\mathbf{c}_{ij}\in \{0,1\}|i=1,2,...,\mathcal{N}_v, j=1,2,...,\mathcal{N}_t\}$ denotes if a visual-text pair matches, where $c_{i j} = 1$ signifies $(\mathbf{v}_i, \mathbf{t}_j)$ is a corresponding visual-text pair, and vice versa. 

Meanwhile, in deep metric learning, it is challenging to optimize every feature to its exact position, generally we leverage a margin $\gamma$ to separate positive and negative pairs. Thus, for typical visual-to-text retrieval, the instinct learning objective is:

\begin{equation}
\begin{aligned}
    \mathcal{O}_{v2t} &:= S(\mathcal{V}, \mathcal{T}_p) - S(\mathcal{V}, \mathcal{T}_n) \geq \mathcal{C}\cdot\gamma,
\end{aligned}
\end{equation}

where $S(\cdot)$ denotes the similarity metric, $\mathcal{T}_p$ and $\mathcal{T}_n$ are the matching pairs and mismatching pairs to the video set.

Since $\mathcal{C}$ is the hard label, i.e., $\mathcal{C}$ can only be either 0 or 1. Therefore for each iteration, the margin between the positive and negative pair becomes: $(\mathbf{c}_p-\mathbf{c}_n)\gamma=\gamma$. Assuming that we use the cosine similarity as the metric, where the matrix production between L2-normalized features represents their similarity, thus the learning objective becomes:

\begin{equation}
\begin{aligned}
    \mathcal{O}_{v2t} &:= S({\mathbf{v}_i}, {\mathbf{t}_j}) - S({\mathbf{v}_i}, {\mathbf{t}_k}) \geq \gamma\\ 
    &:= \gamma-\mathbf{v}_i^T \mathbf{t}_j + \mathbf{v}_i^T \mathbf{t}_k  \leq 0.
\end{aligned}
\end{equation}

Where j and k are samples in positive and negative sets, respectively. That is exactly the learning objective of MI-MM Loss \cite{mi-mm}, a commonly used loss function for text-image retrieval task, Since our task is bidirectional, i.e., we need video-to-text and text-to-video retrieval simultaneously, thus the loss function can be formulated as:

\vspace{0.5mm}
\begin{equation}
    \mathcal{L}=\sum_{(i, j, k) \in \mathcal{N}} \left[\gamma-\mathbf{v}_i^T \mathbf{t}_j+\mathbf{v}_i^T \mathbf{t}_k\right]_+ +\left[\gamma-\mathbf{t}_i^T \mathbf{v}_j+\mathbf{t}_i^T \mathbf{v}_k\right]_+.
\end{equation}

Here, $ [\cdot]_+$ denotes the ReLU function. However, the EK-100 Multi-Instance Retrieval Challenge introduces a correlation matrix $\mathcal{C}=\{\mathbf{c}_{ij}\in [0,1]|i=1,2,...,\mathcal{N}_v, j=1,2,...,\mathcal{N}_t\}$, meaning $\mathcal{C}$ is no longer a hard label, and $\mathbf{c}_{ij}$ could be any value between 0 and 1. To leverage this prior information, the adaptive MI-MM Loss \cite{egovlp} is proposed, formulated as:
\vspace{0.5mm}
\begin{equation}
    \mathcal{L}=\sum_{(i, j, k) \in \mathcal{N}} [\mathbf{c}_{i j}\gamma-\mathbf{v}_i^T \mathbf{t}_j+\mathbf{v}_i^T \mathbf{t}_k]_+ \\+[\mathbf{c}_{i j}\gamma-\mathbf{t}_i^T \mathbf{v}_j+\mathbf{t}_i^T \mathbf{v}_k]_+.
\end{equation}

While the learning objective of adaptive MI-MM Loss is similar to MI-MM Loss, introducing the relevancy matrix $\mathcal{C}$ to the learning objective also means that the correlation between negative pairs, $c_{ik}$, is not always 0. This makes the learning objective less precise for this challenge. Moreover, EgoVLP \cite{egovlp} employs a hard mining strategy that define the positive set as $i^+=\{j|c_{ij}\geq 0.1\}$, i.e., the partially matched video-text pairs could be treated as the positive samples. This can be problematic when $c_{ij} < c_{ik}$, leading the learning objective in the opposite direction to the correct one.

\subsection{Revisit Multi-Similarity Loss}

Meanwhile, Multi-Similarity Loss \cite{ms_loss} demonstrates its effectiveness in metric learning, it is formulated as:

\begin{equation}
    \begin{aligned}
        \mathcal{L}_{MS}=\frac{1}{\mathcal{N}} \sum_{i=1}^\mathcal{N} & \left\{\frac{1}{\alpha} \log \left[1+\sum_{j \in P_i} e^{-\alpha\left(S_{i j}-\gamma\right)}\right]\right. \\
        + & \left.\frac{1}{\beta} \log \left[1+\sum_{k \in N_i} e^{\beta\left(S_{i k}-\gamma\right)}\right]\right\},
    \end{aligned}
\end{equation}

where $\mathcal{P}_i$ and $\mathcal{N}_i$ refer to the positive and negative sets corresponding to i-th video clip, $\alpha$ and $\beta$ are the scale factors for positive and negative pairs, respectively. To simplify this loss function, we consider a special case when $\alpha, \beta \xrightarrow{}\infty$:

\begin{equation}
\mathcal{L}_{MS}^{'}=\sum_{(i, j, k) \in \mathcal{N}} \left[\gamma - \mathbf{S}_{i j}\right]_+ + \left[\mathbf{S}_{i k} -\gamma \right]_+.
\end{equation}

This reveals that the learning objective for Multi-Similarity Loss is to push positive pairs closer to the margin while pulling negative pairs away from it. This inspires us to define a symmetric loss function for positive and negative pairs. However, as previously illustrated, it is challenging to determine if $\mathbf{t}_j$ and $\mathbf{t}_k$ are relatively more positive to the video clip $\mathbf{v}_i$, Therefore, directly applying Multi-Similarity Loss to this challenge is still far from satisfaction.

\subsection{Symmetric Multi-Similarity Loss}

To address the aforementioned issues, we formulate the correlation $\mathcal{R}$ between $S_{ij}$ and $S_{ik}$ as follows:

\begin{equation}
    \mathcal{R} =\sum_{(i, j, k) \in \mathcal{N}} \mathbf{c}_{ij} - \mathbf{c}_{ik}.
\end{equation}

There are three distinct scenarios to consider. When $\mathcal{R}>0$, $S_{ij}$  is the relatively more positive pair compared to $S_{ik}$, and vice versa. Meanwhile, when $\mathcal{R}=0$, the distance between $S_{ij}$ and $S_{ik}$ should be optimized to 0. However, in practice, we find that the loss at $\mathcal{R}=0$ tend to be the dominant loss since the value of $\mathcal{R}$ is very small. To mitigate this, we introduce a relaxation factor, $\tau$, such that when the Euclidean distance between $S_{i j}$ and $S_{i k}$ is smaller than $\tau$, we cease optimizing this part. This adjustment allows us to maintain the major learning objective, i.e., $\mathcal{O}:=S_{p}-S_{n}>\mathcal{R}\gamma$. Thus, we obtain a symmetric loss regarding the distance between positive and negative pairs:

\begin{equation}
    \mathcal{L}_{SMS}=\sum_{(i, j, k) \in \mathcal{N}}\left\{
\begin{array}{lll}
\left[\mathcal{R}\gamma-S_{i j}+S_{i k} \right]_+&  & {\mathcal{R}>0} \\
\left[-\mathcal{R}\gamma+S_{i j}-S_{i k} \right]_+&  & {\mathcal{R}<0} \\
\left[\Vert S_{i j}-S_{i k} \Vert_{1} - \tau \right]_+&  & {\mathcal{R}=0}
\end{array} \right.
\end{equation}

Here, $S_*$ denotes both the similarity of video-to-text and text-to-video.

Theoretically, the relaxation factor $\tau$ should be less than the minimum value of $\mathcal{C}$ for $\mathcal{C}>0$. This ensures that the optimization process remains effective and balanced across different correlation scenarios. However, in practice, we sometimes need a larger $\tau$ to prevent the model from focusing on the similar pairs.



\subsection{Inference Augmentation}

Inspired by \cite{flip}, we employ a flip function during the inference phase. Generally, we could directly calculate the similarity by the matrix production of features from video and text encoders. Here we first obtain the original features, then flip the video feature horizontally, and feed both the features into the model, Finally, we add the obtained features together before calculating the similarity matrix. Pytorch-style pseudo code are shown in Alg. \ref{alg:filp}.

\begin{algorithm}[h]
\caption{Pseudo code for our augmentation method.}
\label{alg:filp}
\begin{lstlisting}[style=python_style]
#Input: V(N,T,C,H,W), T(N,L)
#Output: Similarity Matrix S(N,N)

#Filp the Width of each frame.
V_filp = torch.flip(V, dim=[-1]) 
V_feat, T_feat = model(V,T)
V_feat_flip, T_feat_flip = model(V_flip, T)
V_feat += V_feat_flip
T_feat += T_feat_flip
S = torch.matmul(V_feat,T_feat)
\end{lstlisting}
\end{algorithm}
\section{Experiments}

\subsection{Implementation Details}

We directly utilized the framework, as well as the pretrained models from AVION \cite{AVION}, which is a vanilla  CLIP-based model trained on the LLM-augmented Ego4D dataset \cite{ego4d, lavila}. We then fine-tuned the model using our SMS loss on the EK-100 dataset \cite{ek100_1, ek100_2}. During training, we conduct the experiments on $4\times$ RTX 6000 Ada GPU, and the batch size of our ViT-B-based model is 64 per GPU, resulting in the total batch size of 256. For our ViT-L-based model, we could only fit 60 video clips on every 48GB GPU, resulting in the total batch size of 240. The dimension of each video clip is $16\times3\times224\times224$, indicating that we sample 16 frames per video clip, with each frame resized to a height and width of 224 pixels. We use the AdamW optimizer \cite{adamw} with a learning rate of $2\times10^{-5}$ and train the model for the warmup and total epoch of 1 and 100, respectively. The dimension of feature space is set to 256. For our SMS loss, the margin $\gamma$ is set to 0.6 and relaxation factor $\tau$ is set to 0.1.

\subsection{Ablation Study and Competition Result}

To verify the effectiveness and robustness of our SMS loss, we conduct an ablation study on both our ViT-B-based and ViT-L-based models. The experiment results are presented in Table \ref{tab:result}. Note that we report the first three significant digits without rounding, which may result in slight differences from the public leaderboard. 

All experiments across different loss functions are conducted under the same learning rate and optimizer settings. We use the best-performing hyperparameters for each loss function. Specifically, a margin of 0.2 for the MI-MM loss and 0.4 for the adaptive MI-MM loss.

\begin{table}[h]
    \centering
    \resizebox{1.0\columnwidth}{!}{
    \begin{tabular}{cc|cccccc}
      \toprule[0.5mm]
      &\multirow{2}{*}{\textbf{Methods}}& \multicolumn{3}{c}{\textbf{ mAP (\%)}} & \multicolumn{3}{c}{\textbf{ nDCG (\%)}}\\
      & & V$\rightarrow$ T & T$\rightarrow$ V & Avg. & V$\rightarrow$ T & T$\rightarrow$ V & Avg. \\
      \midrule
      \midrule
      \multirow{4}{*}{\rotatebox{90}{\textit{ViT-B-16}}}
      &MI-MM  & 55.5  & 48.8 & 52.1 & 68.4 & 66.3 & 67.3  \\
      &Adaptive MI-MM & 60.5  & \underline{49.6} & 55.1 & 69.7 & \underline{66.5} & 68.1  \\
      &SMS w/o $\tau$ & \underline{62.2}  & 48.1 & \underline{55.2} & \underline{70.8} & \underline{66.5} & \underline{68.6}  \\
      &SMS (ours) & \textbf{62.9} & \textbf{51.1} & \textbf{57.0}  &\textbf{71.2} & \textbf{67.3} & \textbf{69.2} \\
      \midrule
      \midrule
      \multirow{5}{*}{\rotatebox{90}{\textit{ViT-L-14}}}
      &MI-MM & 58.7  & 52.7 & 55.7 & 71.9 & 69.4 & 70.6  \\
      &Adaptive MI-MM & 65.0  & 54.6 & 59.8 & 73.3 & 70.0 & 71.6  \\
      &SMS (ours) & 67.3 & 56.9  & 62.1 & 74.7 & 71.2 & 73.0\\
      \cmidrule{2-8}
      &w/ Flip & \underline{67.9} & \underline{57.3} & \underline{62.6} &\underline{75.3} & \underline{71.7} & \underline{73.5} \\
      &w/ Ensemble & \textbf{68.7} & \textbf{58.6} & \textbf{63.7}  &\textbf{75.9} &\textbf{72.4} & \textbf{74.2} \\
      \bottomrule[0.5mm]
    \end{tabular}
    }
    \caption{Ablation study result of loss functions and tricks on EK-100 dataset.}
    \label{tab:result}
\end{table}

For both the ViT-B-based and ViT-L-based models, our SMS loss demonstrates superior performance compared to its counterparts. Specifically,  for the ViT-B-based model, our SMS loss improves the average mAP by 1.9\% and the average nDCG by 2.4\% compared to the adaptive MI-MM loss. Similarly, for the ViT-L-based model, our SMS loss also improves the model on the average mAP by 2.3\% and on the average nDCG by 2.4\%. We can also observe a stable improvement after the flip function is applied. Concretely, the performance of our model improves by 0.5\% on both average mAP and nDCG.

Additionally, we conducted an experiment on the ViT-B-based model with $\tau = 0$ to illustrate the necessity of the relaxation factor. We observed that the performance dropped by 1.8\% on average mAP and 0.8\% on average nDCG compared to when $\tau = 0.1$. This highlights the importance of the relaxation factor in achieving optimal performance.

\subsection{Ensemble Strategy}

Table \ref{tab:ensemble} details the settings of the individual models used in our ensemble learning approach. The parameter lr-end represents the minimum learning rate during the training phase, and all the adaptive MI-MM loss-based methods are trained for 100 epochs. We employ a straightforward ensemble strategy by directly summing the similarity matrices obtained from all models. All the models are augmented during inference phase by the flip function.

Due to time constraints, we trained only one SMS-based model ($\tau=0.1$) for 100 epochs. The remaining two models are fine-tuned for 20 epochs based on the initial SMS-based model ($\tau=0.1$). Compared to the single model, our ensemble model outperforms the best individual models by 1.1\% on average mAP and 0.6\% on average nDCG.

\begin{table}[h!]
\centering

 \resizebox{.95\columnwidth}{!}{
\begin{tabular}{c|cccc}
\toprule[0.5mm]
\textbf{Model Name} & \textbf{lr-end} & \textbf{Margin} & \textbf{mAP} & \textbf{nDCG}\\ [0.5ex]
\hline\hline
Adaptive MI-MM\_1   & 1e-6 & 0.4 & 60.1 & 71.9\\
Adaptive MI-MM\_2   & 2e-7 & 0.4 & 59.6 & 71.6\\
Adaptive MI-MM\_3   & 1e-7 & 0.4 & 59.7 & 71.7\\
\hline
SMS $\tau = 0.05$  & 1e-6 & 0.6 & 62.3 & \underline{73.6} \\
SMS $\tau = 0.1$   & 1e-6 & 0.6 & \underline{62.6} & 73.5 \\
SMS $\tau = 0.12$  & 1e-6 & 0.7 & 62.2 & 73.5 \\
\hline
Ensemble Model    & - & - & \textbf{63.7} & \textbf{74.2} \\ 
$\Delta$          & - & - & \textbf{+1.1} & \textbf{+0.6} \\
\bottomrule[0.5mm]
\end{tabular}}
\caption{Corresponding settings and performance of models used in ensemble learning.}
\label{tab:ensemble}
\end{table}

\vspace{-5mm}
\section{Conclusion}

In this report, we explore the loss function in depth to achieve the precise learning objective of EK-100 Multi-Instance Retrieval Challenge. According to the obtained learning objective, we present a novel loss function called Symmetric Multi-Similarity Loss, which addresses the limitations of existing loss functions. Furthermore, we implement a simple trick on the inference phase to further enhance our model's performance. By combining a more accurate loss function, inference tricks, and ensemble learning, our model achieved leading performance among the participants.

\textbf{Limitations}: Due to the time limit,  we are unable to conduct experiments with a wider range of settings, but directly using the same settings as that for adaptive MI-MM loss. However, the gradient of the SMS loss is much smaller than that of the adaptive MI-MM loss since $\mathcal{R}\leq\mathbf{c}_{ij}$. Therefore, using a higher learning rate could yield better results. 

Additionally, SMS loss requires an extra $\mathbf{c}_{ik}$, meaning that a $B\times B$ relevancy matrix for every batch size $B$ is needed, while it is not readily available from the dataloader. Our current approach is to collect it from file during the loss calculation phase, which increases the training time. We plan to address this issue in future work.
{
    \small
    \bibliographystyle{ieeenat_fullname}
    \bibliography{main}
}


\end{document}